# Comparison of VCA and GAEE algorithms for Endmember Extraction

Douglas Winston. R. S.
Informatics Institute (INF)
Federal University of Goiás (UFG)
Goiânia, Brazil
douglas.winston@acm.org

Gustavo T. Laureano
Informatics Institute (INF)
Federal University of Goiás (UFG)
Goiânia, Brazil
gustavo@inf.ufg.br

Celso G. Camilo Jr.
Informatics Institute (INF)
Federal University of Goiás (UFG)
Goiânia, Brazil
celso@inf.ufg.br

*Abstract*—Endmember Extraction is a critical step in hyperspectral image analysis and classification. It is an useful method to decompose a mixed spectrum into a collection of spectra and their corresponding proportions. In this paper, we solve a linear endmember extraction problem as an evolutionary optimization task, maximizing the Simplex Volume in the endmember space. We propose a standard genetic algorithm and a variation with In Vitro Fertilization module (IVFm) to find the best solutions and compare the results with the state-of-art Vertex Component Analysis (VCA) method and the traditional algorithms Pixel Purity Index (PPI) and N-FINDR. The experimental results on real and synthetic hyperspectral data confirms the overcome in performance and accuracy of the proposed approaches over the mentioned algorithms.

## I. INTRODUCTION

THE prior few years have observed considerable advances in hyperspectral imaging where statistical signal processing has performed a central role in pushing algorithm design and development for hyperspectral data exploitation [1]. It has brought the attention of those who come from various disciplinary areas by searching new applications and making associations between remote sensing and other engineering disciplines [2].

The hyperspectral image has matured to a rapidly growing technique in remote sensing image processing due to recent advances in hyperspectral imaging technology. It makes use of hundreds of contiguous spectral bands expanding the capacity of multispectral sensors using dozens of discrete spectral bands [3]. Nevertheless, in real applications, pixels may contain a mixture of spectra resulting from the combination of multiple substances, named endmembers, in varied abundance fractions. In this situation, the diffused energy is a mixing of the endmember spectra [4]. A challenging assignment underlying hyperspectral applications is the spectral unmixing which describes the decomposition of a mixed pixel into a collection of reflectance spectra called endmember signatures, and the corresponding abundance fractions.

Depending on the type of mixing at each pixel, the pixel-mixed model can be separated into two classes, linear mixed model, and nonlinear mixed model [5]. The linear model considers that a mixed pixel is a linear combination of endmember signatures weighted by their correspondent abundance fractions. The linear model is based on the assumption that the number of substances and their reflectance spectra are known. Therefore, with those assumptions, hyperspectral unmixing is a linear problem for which many solutions have been proposed such as Convex Cone Analysis [6], Pixel Purity Index (PPI) [7], Vertex Component Analysis (VCA) [8], N-FINDER (N-FINDR) [9], Simplex Growing Algorithm (SGA) [10], and variations of those above [11][12][13][14].

Pixel purity index (PPI) is a traditional method for endmember extraction. The algorithm determines the endmembers and a pixel purity index by repeatedly projecting data into a set of random unit vectors [15]. Followed by, N-FINDR which finds a simplex of the maximum volume with a given number of vertices. The procedure begins with a random initial selection of pixels. If the volume of the simplex increases with the new pixel, it is accepted as the new endmember [12]. Differently, the Vertex Component Analysis (VCA) is an unsupervised endmember extraction method that projects data onto a orthogonal direction to the subspace spanned by the endmembers already determined, which is based upon the notion that endmembers are simplex vertices and that the affine transformation of a simplex is a simplex as well [8]. The Simplex Growing Algorithm (SGA) is an endmember finding algorithm which grows simplexes with maximal volume one vertex at a time to find new endmembers where Simplex Volumes (SV) are calculated by the determinants of respective endmember matrices [10].

Algorithms based on VCA have low computational complexity and accurate extraction results [16] but there are still several weaknesses in these algorithms. One is the extraction accuracy will be reduced if the real data cannot meet the assumptions of the simplex structure or if there is an absence of pure pixels [17]. Furthermore, these algorithms generated the initial endmembers randomly, which is not an effective way to the initialization and takes a long time to find a desired set of endmembers. In order to solve these problems, an integration of genetic algorithm (GA) and simplex volume maximization is proposed.

In this paper, we formulate the endmember extraction problem as a single objective maximizing the volume of the simplex derived from the endmember matrix. It was believed that the ground-truth endmembers can be located by finding a collection of pixel vectors whose simplex volume is the largest [9]. The evolutionary approach inserts greater flexibility in the convergence of the algorithm and is possible to reach a better set of pure pixels that best represents the data without relying only on the randomness of the selection phase imposed by interactive methods. Experiments on real hyperspectral images show the effectiveness of the proposed endmember extraction method with rather good computational complexity compared to other methods.

The rest of this paper is organized as follows: Section II introduces some background knowledge of linear mixed model, endmember extraction, and evolutionary algorithms. Section III introduces the objective function used, the proposed endmember extractor in details, and metrics used to compare the signatures results. Experimental study of real and synthetic is described in Section IV. Finally, concluding remarks are given in Section V.

## II. BACKGROUND

We will use the following notations throughtout the paper:
- $Y \in \mathrm{R}^L$ represents the observed data matrix. $L$ is the number of spectral bands, and $N$ denotes the number of total pixels in the data matrix.
- $y \in \mathrm{R}^{L \times 1}$ is the measured spectrum of a pixel.
- $M \in \mathrm{R}^L$ is a endmember matrix containing p endmembers.
- $\alpha \in \mathrm{R}^{P \times 1}$ is a vector containing the fractional abundances of the endmembers in the pixel.
- $n \in \mathrm{R}^{L \times 1}$ is a vector collecting the errors affecting the measuresments at each spectral band.

### A. Linear Mixing Model

In the Linear mixing model (LMM), the spectral response of a pixel is a linear combination of all the pure spectral signature of endmembers present in the pixel [18]. When the endmembers are distributed as discrete paths, the LMM is certainly valid and different endmembers do not interfere with each other. Assuming the linear mixing scenario, each observed spectral vector is given by

$$y_i = \sum_{j=1}^{p} m_{ij}\alpha_j + n_j \quad (1)$$

where $y_i$ denotes the measured reflectance of a pixel at spectral band $i$, $m_{ij}$ is the reflectance of the $j^{th}$ endmember at spectral band $i$, $\alpha_j$ stands for the fractional abundance of the $j^{th}$ endmember, and $n_i$ represents noise and modeling errors for the spectral band $i$. If the data has L spectral bands, Eq. (1) can be rewritten as follows:

$$y = M\alpha + n \quad (2)$$

Owing to physical constraints, abundance fractions are non-negative $(\alpha \geq 0)$ and satisfy the positivity constraint $1^T\alpha = 1$, where 1 is a $P \times 1$ vector of ones. Each pixel can be viewed as a vector in an $L$-dimensional Euclidean space, where each channel is assigned to one axis of space. Since $\alpha \in \mathrm{R}^P$, $1^T\alpha = 1$ and $(\alpha \geq 0)$ construct a simplex, then Eq. (2) is also a simplex denotes by $S_x$ where $x \in \mathrm{R}^L$ and $x = M\alpha$.

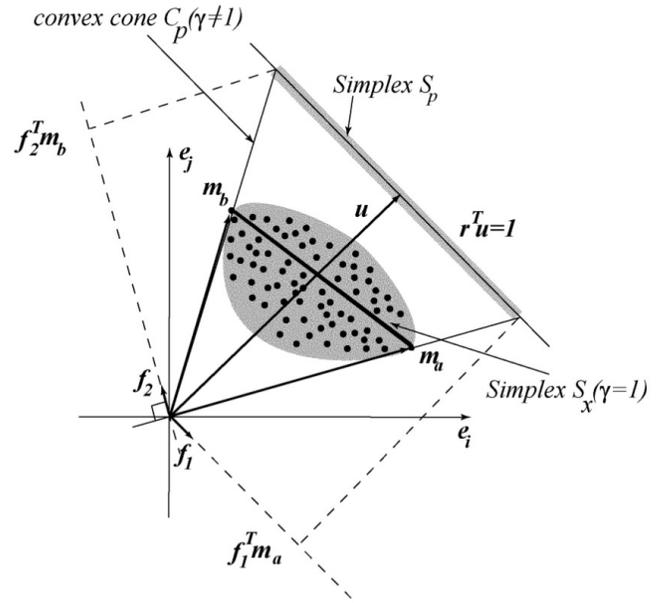

Fig. 1: Illustration of the VCA algorithm [8].

### B. Vertex Component Analysis

The (VCA) algorithm unmix linear mixtures of endmember spectra by assuming $n = 0$ and construction a convex cone $C_p$ following: $y = M\gamma\alpha$, $1^T\alpha = 1$, $(\alpha \geq 0)$ and $(\gamma \geq 0)$ where $\gamma$ is a scale factor to model illumination variability. In Figure 1, the projective projection of the convex cone $C_p$ onto a properly chosen hyperplane is a simplex with vertices corresponding to the vertices of the simplex $S_x$. After identifying $S_p$, the VCA algorithm iteratively projects data onto a direction orthogonal to the subspace spanned by the endmembers already determined. The new endmember signature corresponds to the extreme of the projection [8].

In the first iteration, data are projected onto the first direction $f_1$. The extreme of the projection corresponds to endmember $m_a$. In the next iteration, endmember $m_b$ is found by projecting data onto direction $f_2$, which is orthogonal to $m_a$. The algorithm iterates until the number of endmembers is exhausted.

### C. Genetic Algorithms

Once the genetic representation and the fitness function are defined, a GA proceeds to initialize a population of solutions and then to improve it through repetitive application of the mutation, crossover, and selection operators. Section III-B will explain in details the choice of each operator mentioned and representation of chromosomes in respect to the linear mixing model and the hyperspectral data [19][20].

## D. In Vitro Fertilization Module

The Genetic Algorithm's evolutionary process is composed of a recurrent runs, each repetition produces a new generation. Concerning each fresh generation, individuals are created and injected into the population to succeed some of the current ones, which are dropped. Still, many of these eliminated individuals contain genes with information that is important for the search [21].

In Vitro Fertilization Module (IVFm) recombines the chromosomes from the GA population with new individuals to better exploit information lost. The IVFm generates children from the recombination of chromosome parts of some individuals with the fittest one. If the process generates a better individual, this replaces the current best. Otherwise, if the operator does not produce a better individual, there is no interference in the population [22].

## III. METHODOLOGY

### A. Simplex Volume Maximization

In geometry, a simplex, also named a pentatope, is the generalization of a tetrahedral section of space to arbitrary $n$ dimensions. A $j$-dimensional $(j+1)$-vertex simplex $S_{j+1}$ is a convex polygon with $j+1$ vertices. Therefore, a single point could be considered as a 1-vertex simplex and a 2-vertex simplex could be proclaimed as a line segment between two specified points. Moreover, a 3-vertex simplex is a triangle, a 4-vertex simplex is a tetrahedron, etc [11] [23]. Suppose a $(j+1)$-vertex simplex $S_{j+1}$ where it can be expressed by the set of points as

$$S_{j+1} = S(m_1, m_2, ..., m_{j+1}) \quad (3)$$

The content of volume of a simplex is denoted as $V(S_{j+1})$. Let $V(S_j)$ denote the volume of a $j$-vertex simplex $S_j$ which is a base of a $(j+1)$-vertex simplex $S_{j+1}$ and $h_j$ be the height of the apex from subspace containing the base $S_j$. The volume $V(S_{j+1})$ can be simplified as

$$V(S_{n+1}) = (\frac{1}{n!}) \prod_{j=1}^{n} h_j = (\frac{1}{n!}) V(P_n) \quad (4)$$

By the previous equation it can be seen that the volume of $V(S_n)$ of an n-simplex is $(\frac{1}{n!})$ of the volume of the corresponding $(n-1)$-vertices simplex.

Now, consider an $n$-dimensional simplex $S_n$ with $n+1$ vertices $m_1, m_2, ... m_{n+1} \epsilon R_n$ (endmembers following the hyperspectral representation of a simplex). The volume of $P_n$ is determined by $n$ edge vectors, $M_E = m_2 - m_1, ... m_{n+1} - m_1$, corresponding to $S_n$, it is the absolute value $Det(M_E)$.

$$V(P_n) = |Det(M_E)| = |Det[(m_2 - m_1), ...(m_{n+1} - m_1)]| \quad (5)$$

Therefore, by (4) and (5), the volume $V(S_{n+1})$ of an $(n+1)$-simplex $S_{n+1}$ is expressed as

$$V(S_{n+1}) = (\frac{1}{n!}) V(P_n) = (\frac{1}{n!}) |Det(M_E)| \quad (6)$$

Using elementary column operation of matrix and properties of the determinant for a block matrix, $Det[M_E]$ is obtained by

$$Det \begin{bmatrix} 1 & 1 & ... & 1 \\ m_1 & m_2 & ... & m_{n+1} \end{bmatrix}$$

$$Det[m_2^*, m_3^*, ..., m_{n+1}^*] \quad (7)$$

Then,

$$V(S_{n+1}) = (\frac{1}{n!}) Det[m_2^*, m_3^*, ..., m_{n+1}^*] \quad (8)$$

It is deserving noting that the volume derived by (8) is only possible while $M_E$ is square matrix. Consequently a dimensionality reduction is necessary, in this paper, it was adoted the Principal Component Analysis (PCA) [24].

### B. Genetic Algorithm Endmember Extractor

**Coding Scheme:** In this paper, a chromosome is designed by having $m$ genes where $m$ represents the number of endmembers defined previously by the user or a dimensionality estimation method. To obtain a possible solution to the endmember extraction problem, each gene is considered to be an endmember (pure pixel) in the data, and each individual in the population is considered to be a combination of endmembers (pixels). To generate a coding system, a vector of integer digits in the chromosome is designed where a gene equals to a pixel index in the hyperspectral data. Each endmember contributes to the calculation of a fitness function (Simplex Volume).

**Fitness function:** Under the linear mixing model, the spectral data from a scene resides within a simplex in a multidimensional space whose endmembers represent the vertices of the simplex, and the volume defined by the endmembers is maximal. Therefore, the objective is to find the maximum volume among the set of all simplexes with the same dimension. The fitness function then is the calculation of the volume of a simplex expressed by the Eq. (8)

**Initialization:** Population individuals are created by a randomly process where arbitrary pixels are chosen from the data matrix as individual's genes.

**Selection:** Tournament selection is used to select three individuals at random from the population, each individual selected goes through the mating process. [20].

**Mating:** A two-point crossover is chosen for the mating process of individuals. The two individuals are modified in place and both keep their original length depending on a probability informed at the beginning of the evolutionary process.

**Mutation:** From a selected individual, randomly choose a new gene (Pixel) from the data matrix replacing the previous gene. The gene alteration depends on a probability defined at the inicialization of the GA paramenters.

**In Vitro Fertilization Module:** At the beginning of the GA execution the same population selected by the GA is then inject into the IVF module where each generation executes the following tasks [21]:

1) Find the fittest individual and label it as Father;

2) Receive as input the following parameters: the N GA individuals (NuIndiv), which are handled in the recombination process;
3) From N selected individuals, the half (N / 2) changes the chromosomes, forming N';
4) The fittest (Father) and the N' individuals are recombined. As there is only one father for each recombination process, the generated children are siblings;
5) Each individual from the IVF module is the compared to the population, if it generates good individuals, they are inserted in the new population, replacing elected individuals chosen by the elitism process. Without elitism, individuals replace any others aleatory.

**VCA Extension:** From GAEE and GAEE-IVFm populations one individual is substituted by a VCA solution with the aim to increase convergence rates and possibly get a better local solution than VCA.

### C. Spectral Angle Mapping

To determine the spectral similarity between given reference spectra,$t$,and the spectra estimated by the endmember extractors,$r$, the spectral angle mapper (SAM) is used [25]. The SAM result is reported as the angular difference (in radians) between two spectra according to the equation:

$$\cos^{-1} \frac{t \cdot r}{\|t\| \cdot \|r\|} \quad (9)$$

which is equal to,

$$SAM(t,r) = \cos^{-1} \frac{\sum_{i=1}^{P} t_i \cdot r_i}{\sum_{i=1}^{P} t_i^2 \sum_{i=1}^{P} r_i^2} \quad (10)$$

### D. Spectral Information Divergence

Comparable to the Spectral Angle Mapper (SAM), which has been widely used in the past, Spectral Information Divergence (SID) is other effective metric that preservers spectral properties while comparing signatures. SID is based on a spectral similarity measure to capture the spectral correlation between two pixels [26]. Assume $y = (y_1, ..., y_L)^T$ is a pixel with probability $q = (q_1, ..., q_L)^T$ where $q_j = y_i / \sum l = 1^L y_l$. We can define the $l$-th band self-information of x and y as follows :

$$I_l(x) = -\log p_t$$
$$I_l(y) = -\log q_t \quad (11)$$

Using (11), the relative entropy of $y$ with respect to $x$ is defined by

$$D(x\|y) = \sum_{l=1}^{L} p_l D(x\|y)$$
$$= \sum_{l=1}^{L} p_l (I_l(y) - I_l(x)) \quad (12)$$
$$= \sum_{l=1}^{L} p_l \log\left(\frac{p_l}{q_l}\right)$$

Following (12), we can define a symmetric hyperspectral measure, known as spectral information divergence (SID) by

$$SID(x,y) = D(x\|y) + D(y\|x) \quad (13)$$

## IV. EXPERIMENTAL RESULTS

In this section, we compare PPI, N-FINDR, VCA, the proposed GAEE version and GAEE with IVF module (IVFm). The algorithms run on a processor Intel Core i7 4750HQ 2.0 GHz, 8GB of RAM, Windows 10, and Python using the libraries *numpy* and *sklearn* as a resource to mathematical representation and *deap* [27] for genetic algorithm construction and evolution.

The experiments are conducted aiming evaluate the algorithms by the metrics $SAM$, $SID$ and the running time using a real and synthetic data (described below). Since all algorithms contains a stochastic factor, the experimental results follow 50 Monte Carlo runs. For the proposed algorithms, GAEE and GAEE-IVFm, it was selected the best parameters values for $p_m = (0.05, 0.1, 0.3)$, $p_c = (0.5, 0.7, 1)$ with $n_{pop} = 100$ and $n_{gen} = 1000$, corresponding mutation probability, crossing probability, population size and number limit of generations, respectively. Table I presents the selected configuration for each version of GAEE algorithm which results is presented in Tables V and VII. We also present in Tables VI and VIII a variance t-test for p-values over 0.05 of significance to support and measure algorithm robustness.

TABLE I: Best Parameters selected for each GAEEs algorithm base on the 50 Monte Carlo runs

|  | $p_m$ | $p_c$ |
|---|---|---|
| *GAEE* | 0.1 | 1 |
| *GAEE-IVFm* | 0.3 | 0.7 |
| *GAEE-VCA* | 0.05 | 0.5 |
| *GAEE-IVFm-VCA* | 0.1 | 1 |

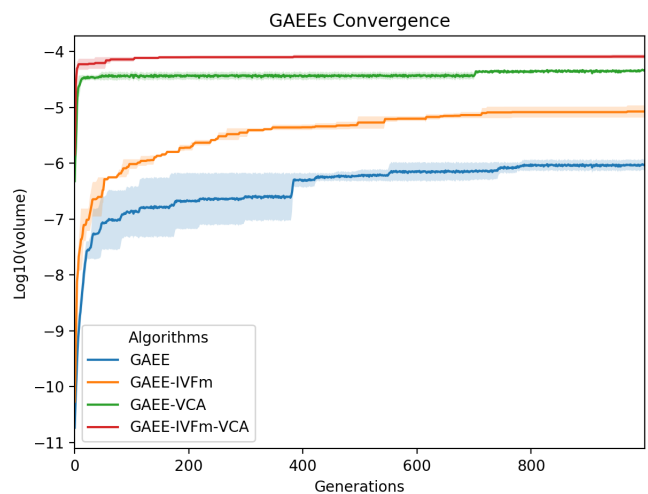

Fig. 2: GAEEs Convergence rate through 1000 generations and 50 Monte Carlo runs using **Cuprite Dataset**

## A. Cuprite Dataset

Cuprite is the most benchmark dataset for the hyperspectral unmixing research that covers the Cuprite in Las Vegas, NV, U.S. There are 224 channels, ranging from $370nm$ to $2480nm$. After removing the noisy channels (1–2 and 221–224) and water absorption channels (104–113 and 148–167), we remain 188 channels. A region of $250 \times 190$ pixels is considered, where there are 14 types of minerals. Since there are minor differences between variants of similar minerals, the number of endmembers is reduce to 12, which are summarized as follows "#1 Alunite", "#2 Andradite", "#3 Buddingtonite", "#4 Dumortierite", "#5 Kaolinite1", "#6 Kaolinite2", "#7 Muscovite", "#8 Montmorillonite", "#9 Nontronite", "#10 Pyrope", "#11 Sphene", "#12 Chalcedony".

The Tables V and VII show the mean of SAM and SID distances for the Cuprite Dataset after 50 runs. From these results, we can observe that GA based algorithms overcame VCA, PPI and N-FINDR results in most of the extracted endmembers, with advantage for the GAEE algorithm. The GAEE-IVFm tends to approximate to GAEE achieving best results in some endmembers. The addition of endmember hint in GAEE-VCA and GAEE-IVFm-VCA does not contribute significantly to achieve best results compared with GAEE algorithm but approximate the metrics with a considerable reduction in population size and number of generations, such as the GAEE-IVFm algorithm. Figure 2 shows GAEE high diversity of individuals at early 200 generations while the extended VCA versions induce fast convergence but low diversity. The last row of tables presents the Root Mean Squared (RMS) for both SAM and SID metrics.

In Tables VI and VIII we noticed that GAEE had $7.92\%$ gain over VCA with SAM metric and $24.37\%$ gain with SID metric, but the VCA performs much faster than the evolutionary approaches. In Figure 3, the 12 extracted endmembers are shown for the interactive algorithms and the GAEE algorithm version with lowest RMS (GAEE-IVFm).

## B. Legendre Dataset

Legendre Dataset is a set of hyperspectral synthetic images from the IC Synthetic Hyperspectral Collection. These images were constructed by the "Synthesis tools" package [28]. All these synthetic images have been generated using five selected endmembers from the USGS spectral library included in the "Synthesis tools" package. Each image's spatial dimensions are of $128$ pixels and they have $431$ spectral bands. In this paper three additive noise versions of the legendre dataset has been used, with Signal to Noise Ratios (SNR) of 20, 40 and 80 dB, respectively.

Tables II and III show both SAM and SID distances for the three synthetic images generated by the endmember extractors. It can be noticed that GAEE-IVF had the best results among all algorithms except in the image with low SNR (20db) due to option of not using a dimentionality reduction or noise filter on the evolutionary process. Other evolutionary versions performed similar to GAEE-IVF and better than VCA in high SNR. Table IV presents the SAM gain over the methods used in the comparison. In the case where the SNR is 40 dB, the GAEE-IVF has 84% of gain and 114% over images with SNR of 80 dB. In Figure 4a and 4b is possible to notice that the GAEE-IVF extracted the right pure pixel as endmember, but the additive noise contributed to the low SAM distance. However, the SID distances were best for the evolutionary algorithm GAEE-IVF with a VCA initial individual's solution is added to the GAEEs population.

TABLE II: Root Mean Squares **SAM** Distance Between Extracted Endmembers and Laboratory Reflectances for **Legendre Dataset** with **Signal Noise Ratio** of 20, 40 and 80 dB

|  | **20 dB** | **40 dB** | **80 dB** |
|---|---|---|---|
| *PPI* | 0.3348 | 0.2474 | 0.1281 |
| *N-FINDR* | 0.1312 | 0.0685 | 0.0691 |
| *VCA* | **0.0851** | 0.0660 | 0.0691 |
| *GAEE* | 0.0924 | 0.0424 | 0.0412 |
| *GAEE-IVFm* | 0.0939 | **0.0375** | **0.0354** |
| *GAEE-VCA* | 0.1212 | 0.0559 | 0.0691 |
| *GAEE-IVFm-VCA* | 0.0971 | 0.0544 | 0.0496 |

TABLE III: Root Mean Squares **SID** Distance Between Extracted Endmembers and Laboratory Reflectances for **Legendre Dataset** with **Signal Noise Ratio** of 20, 40 and 80 dB

|  | **20 dB** | **40 dB** | **80 dB** |
|---|---|---|---|
| *PPI* | 34.9975 | 19.8908 | 6.3210 |
| *N-FINDR* | 3.3918 | 7.7910 | 8.2932 |
| *VCA* | 2.8671 | 7.5158 | 8.2847 |
| *GAEE* | 2.4536 | 3.7487 | 4.0112 |
| *GAEE-IVFm* | 2.5987 | **1.4021** | **1.4814** |
| *GAEE-VCA* | 2.5749 | 6.7584 | 8.2933 |
| *GAEE-IVFm-VCA* | **1.6756** | 2.6890 | 3.6275 |

TABLE IV: GAEE-IVFm SAM Gain Comparison for **Legendre Dataset** with **Signal Noise Ratio** of 20, 40 and 80 dBs

|  | **20 dB** | **40 dB** | **80 dB** |
|---|---|---|---|
| *PPI* | 298.45% | 988.78% | 454.87% |
| *N-FINDR* | 26.11% | 96.26% | 114.40% |
| *VCA* | **-42.25%** | 84.86% | 114.39% |
| *GAEE* | -0.43% | 17.18% | 22.16% |
| *GAEE-IVFm* | - | - | - |
| *GAEE-VCA* | 16.81% | 78.16% | 114.40% |
| GAEE-IVFm-VCA | 3.22% | 33.26% | 19.50% |

## V. CONCLUSION

This paper discussed the possible application of an evolutionary algorithm (GAEE) [1] and its variation (GAEE-IVFm) to the problem of finding a set of endmembers with maximal volume in a hyperspectral image. First, from the concepts detailed by the Vertex Component Analysis and the Linear Mixing Model it was possible to construct a genetic model of the hyperspectral data, setting a simplex representation for the datasets and performing a selection of pure pixels in each generation until a limit is reached. The results of GAEE were superior to the state-of-art method VCA with 7.92% gain for the Cuprite dataset with SAM metric and over 84% gain for the synthetic data with SNR of 40 dB and 114% of gain for 80 dB. GAEE-IVFm was considered a superior method while working with synthetic images, since the presence of pure pixels (endmembers) and the linear mixing model are guaranteed by its construction. The methods developed in this paper obtained inferior results in the synthetic dataset with SNR of 20 db. The inferior results are due to the fact that the evolutionary approaches do not make use of projections or dimentionality reduction, differently from VCA, which reduces the noise factor. An immediate next step for this research is to develop a noise filter as post processing of the results for images with low Signal Noise Ration. In Addition, the linear mixing model could be replaced by a non-linear kernel, then eliminating the necessity of pure pixels on the image. Consequently, the choice of a new evolutionary fitness function would reduce the computational complexity improving GAEEs performance.

---

[1]The proprosed methods are available at https://github.com/haruto-boshi/GAEE

TABLE V: Comparison between the ground-truth Laboratory Reflectances and extracted endmembers using PPI, N-FINDR, VCA, GAEE, GAEE-IVFm and the variations GAEE-VCA and GAEE-IVFm-VCA using **SAM** metric for the **Cuprite Dataset**.

|  | PPI | N-FINDR | VCA | GAEE | GAEE-IVFm | GAEE-VCA | GAEE-IVFm-VCA |
|---|---|---|---|---|---|---|---|
| *Alunite* | 0.3744 | 0.1122 | **0.0939** | 0.1122 | 0.1034 | 0.1043 | 0.1043 |
| *Andradite* | 0.0758 | 0.2068 | 0.1034 | **0.0693** | 0.0760 | 0.1694 | 0.1694 |
| *Buddingtonite* | 0.2081 | 0.1205 | 0.0786 | 0.0798 | **0.0762** | 0.0762 | 0.0762 |
| *Dumortierite* | 0.1907 | 0.0706 | **0.0702** | 0.0735 | 0.0719 | 0.0755 | 0.0755 |
| *Kaolinite_1* | **0.0795** | 0.0870 | 0.0862 | 0.0952 | 0.0935 | 0.0870 | 0.0870 |
| *Kaolinite_2* | 0.0820 | 0.0992 | 0.0741 | **0.0649** | 0.0723 | 0.0744 | 0.0782 |
| *Muscovite* | 0.2506 | 0.0961 | 0.1805 | **0.0861** | 0.1091 | 0.0965 | 0.0961 |
| *Montmorillonite* | 0.1338 | **0.0646** | 0.0651 | 0.0671 | 0.0677 | 0.0688 | 0.0650 |
| *Nontronite* | 0.1033 | 0.0780 | 0.0801 | **0.0711** | 0.0791 | 0.1150 | 0.1150 |
| *Pyrope* | 0.0579 | 0.0865 | 0.0818 | **0.0563** | 0.0623 | 0.0793 | 0.0686 |
| *Sphene* | 0.0673 | 0.0542 | **0.0530** | 0.1121 | 0.0946 | 0.0795 | 0.0901 |
| *Chalcedony* | 0.0871 | **0.0731** | 0.0773 | 0.0738 | 0.0756 | 0.0765 | 0.0861 |
| *RMS* | 0.1316 | 0.0884 | 0.0803 | **0.0740** | 0.0755 | 0.0848 | 0.0855 |

TABLE VI: Statistics and Performance measurements for the 50 Monte Carlo runs comparing PPI, N-FINDR, VCA, GAEE, GAEE-IVFm and the variations GAEE-VCA and GAEE-IVFm-VCA using **Cuprite Dataset** and **SAM** metric.

|  | PPI | N-FINDR | VCA | GAEE | GAEE-IVFm | GAEE-VCA | GAEE-IVFm-VCA |
|---|---|---|---|---|---|---|---|
| *Mean* | 0.1425 | 0.1033 | 0.1024 | 0.1016 | **0.0989** | 0.1109 | 0.1090 |
| *Std* | **0.0000** | 0.0225 | 0.0252 | 0.0248 | 0.0255 | 0.0117 | 0.0157 |
| *P-value* | -34.8557 | -0.6562 | 0.0000 | 0.5032 | 2.3017 | -6.4153 | -4.8551 |
| *Gain* | 38.95% | 16.31% | 7.92% | 0% | 2.06% | 12.79% | 13.50% |
| *Time* | 2.1929 | 7.8318 | **0.5106** | 8.9232 | 22.3329 | 8.7494 | 22.1761 |

TABLE VII: Comparison between the ground-truth Laboratory Reflectances and extracted endmembers using PPI, N-FINDR, VCA, GAEE, GAEE-IVFm and the variations GAEE-VCA and GAEE-IVFm-VCA using **SID** metric for the **Cuprite Dataset**.

|  | PPI | N-FINDR | VCA | GAEE | GAEE-IVFm | GAEE-VCA | GAEE-IVFm-VCA |
|---|---|---|---|---|---|---|---|
| *Alunite* | **0.0000** | **0.0000** | 0.0105 | 0.0170 | **0.0000** | **0.0000** | 0.0145 |
| *Andradite* | **0.0000** | 0.0117 | 0.0052 | 0.0055 | 0.0092 | 0.0077 | 0.0056 |
| *Buddingtonite* | 0.0477 | 0.0196 | 0.0077 | 0.0076 | 0.0108 | **0.0072** | 0.0072 |
| *Dumortierite* | 0.0562 | **0.0071** | 0.0298 | 0.0072 | 0.0181 | 0.0077 | 0.0077 |
| *Kaolinite_1* | 0.0114 | **0.0104** | 0.0139 | 0.0139 | 0.0128 | 0.0131 | 0.0131 |
| *Kaolinite_2* | 0.0114 | 0.0058 | 0.0042 | 0.0049 | **0.0029** | 0.0111 | 0.0086 |
| *Muscovite* | 0.0969 | 0.0317 | 0.0148 | **0.0086** | 0.0286 | 0.0285 | 0.0171 |
| *Montmorillonite* | 0.0230 | 0.0053 | **0.0047** | 0.0052 | 0.0048 | 0.0057 | 0.0060 |
| *Nontronite* | 0.0126 | 0.0083 | 0.0093 | **0.0065** | 0.0082 | 0.0155 | 0.0155 |
| *Pyrope* | 0.0071 | 0.0438 | 0.0229 | **0.0057** | 0.0279 | 0.0593 | 0.0593 |
| *Sphene* | 0.0076 | 0.0912 | 0.0096 | 0.0165 | 0.0086 | 0.0099 | **0.0067** |
| *Chalcedony* | 0.0088 | 0.0093 | **0.0069** | 0.0069 | 0.0096 | 0.0070 | 0.0070 |
| *RMS* | 0.0217 | 0.0188 | 0.0107 | **0.0081** | 0.0109 | 0.0133 | 0.0129 |

TABLE VIII: Statistics and Performance measurements for the 50 Monte Carlo runs comparing PPI, N-FINDR, VCA, GAEE, GAEE-IVFm and the variations GAEE-VCA and GAEE-IVFm-VCA using **Cuprite Dataset** and **SID** metric.

|  | PPI | N-FINDR | VCA | GAEE | GAEE-IVFm | GAEE-VCA | GAEE-IVFm-VCA |
|---|---|---|---|---|---|---|---|
| *Mean* | 0.0236 | 0.0257 | 0.0197 | **0.0163** | 0.0194 | 0.0265 | 0.0268 |
| *Std* | **0.0000** | 0.0064 | 0.0107 | 0.0099 | 0.0097 | 0.0065 | 0.0059 |
| *P-value* | -5.2271 | -7.2962 | 0.0000 | 3.8716 | -7.4272 | -8.2866 | **0.0752** |
| *Gain* | 50.65% | 56.79% | 24.37% | 0 | 25.44% | 38.91% | 37.31% |

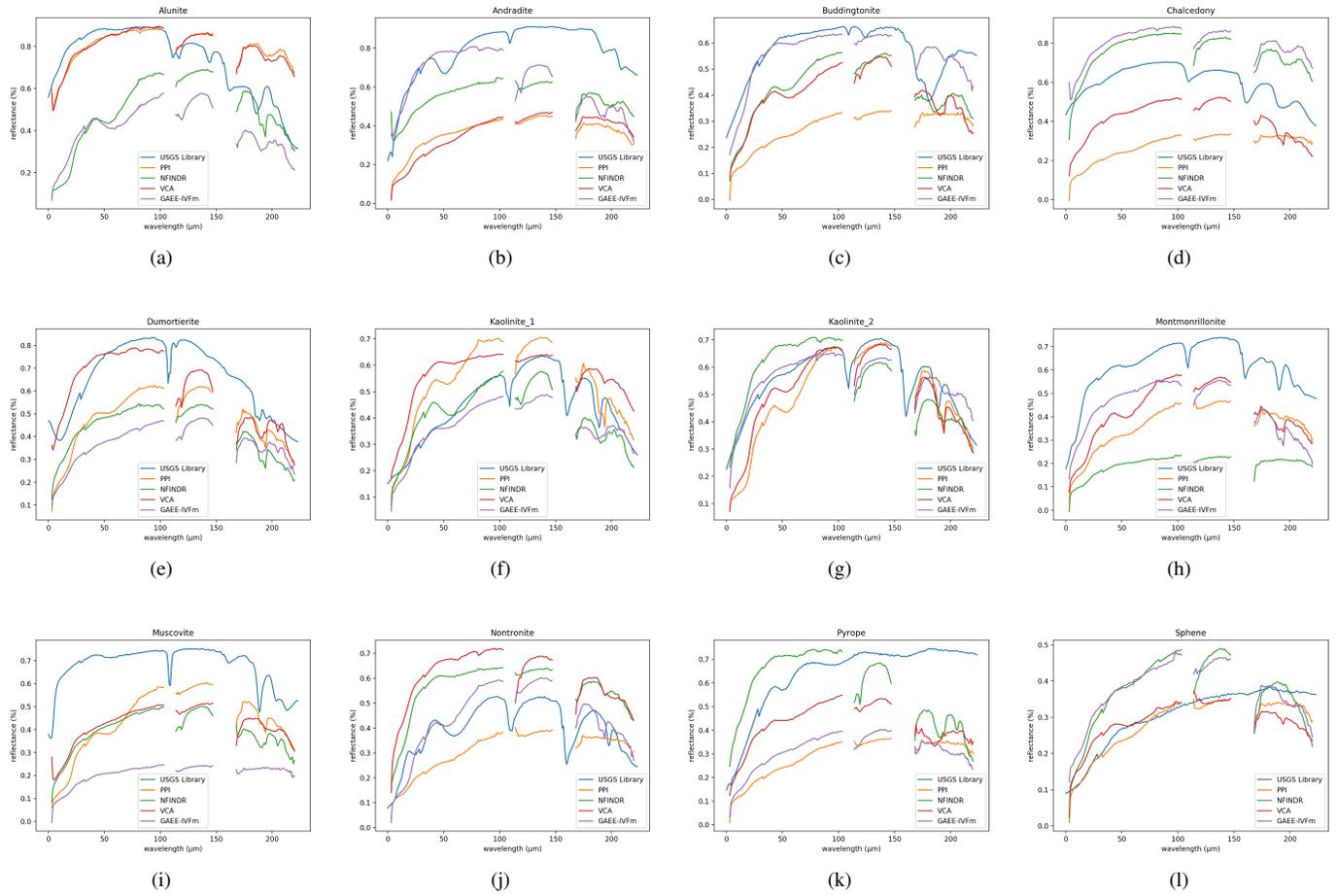

Fig. 3: Extracted Signatures Comparison Between **VCA**, **N-FINDR**, **GAEE-IVFm** and The USGS Spectral Library for the **Cuprite Dataset**

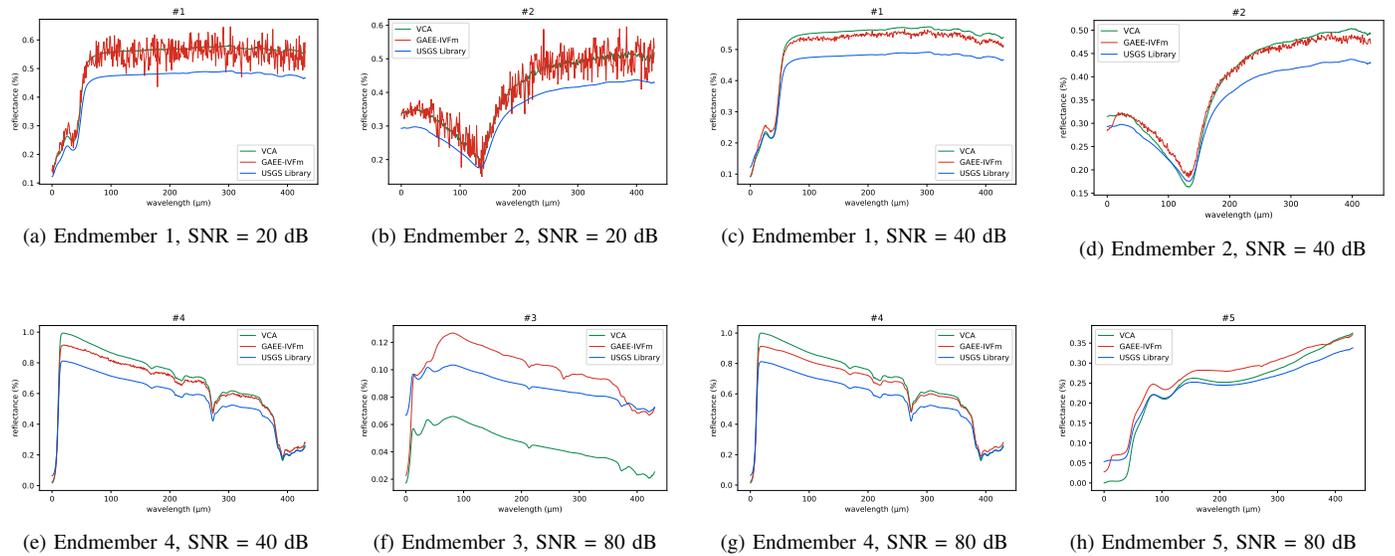

Fig. 4: Extracted Signatures Comparison Between **VCA**, **GAEE-IVFm** and The USGS Spectral Library for the **Legendre Dataset**